\documentclass{article}

\usepackage[preprint]{neurips_2025}

\usepackage{booktabs}
\usepackage{multirow}
\usepackage[table]{xcolor}

\usepackage[utf8]{inputenc} 
\usepackage[T1]{fontenc}    
\usepackage{hyperref}       
\usepackage{url}            
\usepackage{booktabs}       
\usepackage{amsfonts}       
\usepackage{graphicx}       
\usepackage{nicefrac}       
\usepackage{microtype}      
\usepackage{xcolor}         
\usepackage{wrapfig}
\usepackage{listings}

\title{LightRouter: Towards Efficient LLM Collaboration with Minimal Overhead}

\author{
Yifan Zhang$^{1}$,
Xinkui Zhao$^{1*}$,
Zuxin Wang$^{1}$,
Guanjie Cheng$^{2}$,
Yueshen Xu$^{3}$,\\
\textbf{Shuiguang Deng$^{2}$,
Jianwei Yin$^{2}$} \\
$^1$School of Software Technology, Zhejiang University, Hangzhou, China \\
$^2$School of Computer Science, Zhejiang University, Hangzhou, China \\
$^3$School of Software Engineering, Xidian University, Xi'an, China \\
$^*$Corresponding author.
}
\newcommand{\text}[1]{\textrm{#1}}

\begin{document}

\maketitle

\begin{abstract}
The rapid advancement of large language models has unlocked remarkable capabilities across a diverse array of natural language processing tasks. However, the considerable differences among available LLMs—in terms of cost, performance, and computational demands—pose significant challenges for users aiming to identify the most suitable model for specific tasks. In this work, we present LightRouter, a novel framework designed to systematically select and integrate a small subset of LLMs from a larger pool, with the objective of jointly optimizing both task performance and cost efficiency. LightRouter leverages an adaptive selection mechanism to identify models that require only a minimal number of boot tokens, thereby reducing costs, and further employs an effective integration strategy to combine their outputs. Extensive experiments across multiple benchmarks demonstrate that LightRouter matches or outperforms widely-used ensemble baselines, achieving up to a 25\% improvement in accuracy. Compared with leading high-performing models, LightRouter achieves comparable performance while reducing inference costs by up to 27\%. Importantly, our framework operates without any prior knowledge of individual models and relies exclusively on inexpensive, lightweight models. This work introduces a practical approach for efficient LLM selection and provides valuable insights into optimal strategies for model combination.
\end{abstract}

\section{Introduction}
Recent advances in large language models (LLMs; ~\cite{zhang2022opt,chowdhery2023palm, touvron2023llama, brown2020language, achiam2023gpt, liu2024deepseek, guo2025deepseek}) have demonstrated remarkable capabilities across a wide range of natural language understanding and generation tasks, with different LLMs often excelling in distinct areas and outperforming each other on specific benchmarks or applications (~\cite{bi2023oceangpt,ren2025deepseek,zhu2024deepseek}). In addition to their varied capabilities, LLMs also differ significantly in terms of cost, which adds another layer of complexity to model selection. As the landscape of LLMs continues to expand, users face increasing challenges in identifying the most suitable model for a given task, balancing both performance and cost considerations (~\cite{snell2024scaling, chen2024role}). Moreover, efficiently leveraging the collective expertise of multiple LLMs remains an exciting but largely unresolved direction. This situation raises a fundamental question: \textit{How can we effectively select or utilize LLMs to achieve the best possible solution?}

Existing strategies for model selection generally fall into two categories. The first category is to pre-select a single LLM that is theoretically optimal for a given task, based on prior knowledge, domain intuition, or historical performance statistics (~\cite{feng2024graphrouter,xia2024llm}). This approach is appealing for its cost-effectiveness, as it avoids redundant computation by relying on a single model. However, the performance of a single LLM can fluctuate significantly even within the same task, resulting in unstable or suboptimal outputs in practice. Additionally, this method requires continuous monitoring and updating of model knowledge and performance benchmarks, which can be both labor-intensive and impractical as new models and tasks emerge. The second approach involves aggregating the outputs of multiple LLMs using techniques such as majority voting (~\cite{jiang2023llm,ravaut2022summareranker}), averaging, or other ensemble methods (~\cite{wang2024mixture}). This strategy often results in improved answer quality and greater robustness, as it leverages the complementary strengths of different models. However, these benefits come at the expense of significantly increased computational resources and latency, since multiple models must be queried for each task. For example, our experiments show that the popular ensemble method MoA (~\cite{wang2024mixture}) incurs a computational cost exceeding 600\% of that of a single model. Such resource overhead can quickly become prohibitive, particularly in scenarios requiring large-scale deployment.

In this work, we seek a more cost-effective and high-performance approach to harnessing the collective capabilities of multiple LLMs. To this end, we introduce \textbf{LightRouter}---a truly \textbf{out-of-the-box, plug-and-play} framework that balances cost and performance without requiring any prior knowledge of tasks, model ability assessments, or additional training. Our design is inspired by modern distributed systems such as Kubernetes, which employ a two-stage scheduling strategy: they first gather lightweight status information from all candidate workloads through efficient probing (e.g., GET requests), and then judiciously allocate resources to the most promising candidates based on this preliminary assessment. Similarly, LightRouter initially queries all candidate models for a small number of boot tokens, thereby quickly and inexpensively assessing the potential of each model without incurring the full computational cost. Based on these initial outputs, LightRouter filters out models that do not meet quality standards and allows only the most promising models to continue generating responses. Finally, the outputs from the remaining models are ensembled to produce the final, optimal answer. We evaluate LightRouter across several popular benchmarks. Extensive experiments demonstrate that LightRouter matches or surpasses leading ensemble methods, achieving up to a 25\% improvement in accuracy, and delivers performance on par with state-of-the-art (SOTA), high-cost models while lowering inference costs by 27\%. Notably, our framework requires no prior knowledge about individual models and operates efficiently using only inexpensive, lightweight models.
Accordingly, our contributions are as follows:
\begin{itemize}
    \item \textbf{We propose LightRouter, a novel framework for cost-effective and adaptive multi-LLM utilization without requiring prior knowledge of task or model specifics.} Unlike static selection or traditional ensembles, LightRouter dynamically queries models and halts underperformers early, minimizing computational overhead.
    
    \item \textbf{We design a two-stage adaptive selection mechanism:} preliminary partial outputs are used to filter candidates, and only the most promising models proceed to full generation, which significantly reduces unnecessary computation while preserving diversity and robustness.
    
    \item \textbf{We introduce an efficient ensembling strategy that aggregates only the top-performing models,} achieving a strong balance between accuracy and efficiency and avoiding the high costs of full ensembling.
    
    \item \textbf{Extensive experiments on standard LLM benchmarks show that LightRouter achieves higher answer quality and lower inference cost,} and further illuminate why adaptive, selective routing is essential for robust multi-model orchestration.
\end{itemize}

\section{Method}
\subsection{Theoretical Model for LightRouter}
We introduce the theoretical model for LightRouter, which aims to achieve optimal performance-cost trade-offs by leveraging multiple LLMs in a collaborative manner.

\subsubsection{Background}
Given input $X$, a generative model produces $Y = (y_1, \ldots, y_T)$, with $P(Y|X) = \prod_{t=1}^T P(y_t|y_{<t}, X)$. Early errors in autoregressive generation can propagate, degrading later outputs. We quantify semantic consistency with the ground-truth $Y^{\text{true}}$ using $S(y_i, y_i^{\text{true}}) \in [0, 1]$, and model cumulative error up to step $m$ as $\epsilon = 1 - \prod_{i=1}^m S(y_i, y_i^{\text{true}})$.

Suppose $N$ models $\{f_1, \ldots, f_N\}$ generate outputs $Y^{(f_i)}$ with sequence-level consistency $S^{(f_i)}$. Our objective is to maximize $S(Y^{\text{final}}, Y^{\text{true}}) - \lambda T_{\text{total}}$, where $T_{\text{total}}$ is total token cost and $\lambda$ is a trade-off parameter.

\subsubsection{LightRouter Method}
\paragraph{Merging Multiple Outputs}
Given $N$ models ${f_1, \dots, f_N}$, each generates an output $Y^{(f_i)}$. We aggregate a selected set of outputs using a large model $f_{\text{merge}}$ as $Y^{\text{final}} = f_{\text{merge}}({ Y^{(f_i)} : f_i \in \mathcal{S} })$, where $\mathcal{S}$ is a subset of candidate models. If each $S^{(f_i)}$ is an independent random variable with mean $\mu$ and variance $\sigma^2$, then merging $k$ outputs by $f_{\text{merge}}$ reduces variance to $\sigma^2/k$. This process leverages the collective wisdom of multiple models, leading to higher and more stable semantic consistency in $Y^{\text{final}}$.

\paragraph{Total Cost}
The total token cost is $T_{\text{total}} = \sum_{f_i \in \mathcal{S}} T^{(f_i)} + T^{(\text{merge})}$, where $T^{(f_i)}$ is the token count for model $f_i$’s output, and $T^{(\text{merge})}$ is the cost for the merging step. Larger $\mathcal{S}$ means higher cost.

\paragraph{Top-$k$ Selection}
To maximize the merged output quality while controlling cost, we select the top-$k$ candidates with the highest semantic consistency scores, i.e., $    \mathcal{S}_{\text{top-}k} = \mathop{\mathrm{arg\,top}_k} \left\{ S^{(f_i)} : i = 1, \dots, N \right\}$. If low-quality outputs are included in $\mathcal{S}$, they can degrade the merged result, since $S(Y^{\text{final}}, Y^{\text{true}}) \leq \frac{1}{k} \sum{i=1}^k S^{(f{i})}$, and adding noisy candidates increases risk. The optimal $k^*$ balances semantic consistency improvement and cost, i.e., $k^* = \arg\max_{k \leq N} { \mathbb{E}[S_k] - \lambda T_{\text{total}}(k) }$, where $\mathbb{E}[S_k]$ is the expected consistency after merging $k$ top candidates. Thus, top-$k$ selection ensures only the most semantically aligned outputs are merged, maximizing final performance under budget constraints.

\subsection{Architecture}
\begin{figure}[htbp]
    \centering
    \includegraphics[width=1\linewidth]{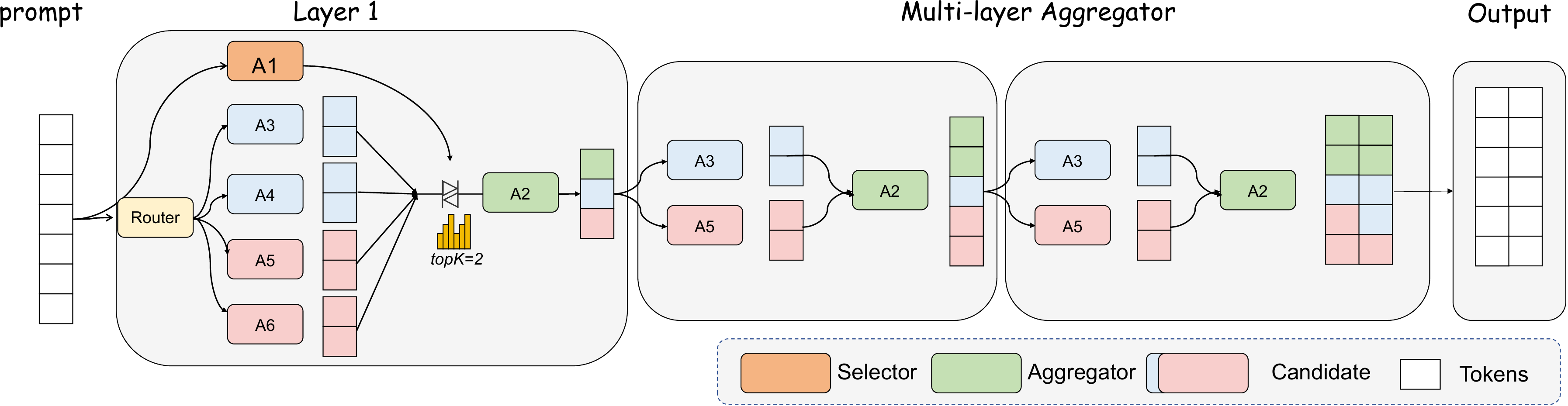}
    \caption{\textbf{Illustration of the LightRouter architecture.} The framework routes prompts through a selector agent and multiple layers of aggregator agents, progressively combining outputs from a pool of candidate models to generate the final response. Different agent roles are color-coded.}
    \label{fig:struc}
\end{figure}

An overview of the LightRouter architecture is depicted in Figure~\ref{fig:struc}. LightRouter is organized as a multi-layer expert-routing system with $l$ layers. The first layer consists of $n$ expert \textbf{Candidate Models}, denoted as $\{f_1, f_2, \dots, f_n\}$, each potentially equipped with specialized capabilities such as coding or mathematical reasoning. For each incoming query $X$, the \textbf{Router} module dispatches the query to each candidate model. Most models are evaluated unless the input requires specific skills, in which case only specialized models are activated.

LightRouter’s scheduling strategy draws inspiration from systems like Kubernetes: rather than fully executing every candidate, it first queries all models for short, preliminary outputs, analogous to Kubernetes’ use of lightweight GET requests for status checking. This enables a rapid, low-cost assessment of each model’s potential.

Each candidate model $f_i$ generates a short initial response $Y^{(f_i)}_0$, which is then evaluated by a \textbf{Selector Model} that assigns a semantic consistency score $S^{(f_i)}$ by comparing outputs. Without access to ground-truth answers, the Selector leverages mutual information among candidate responses to identify those most likely to be correct. It ranks and selects the top-$k$ candidates to proceed, similar to how Kubernetes schedules the most promising workloads for resource allocation, balancing output quality and efficiency.

The selected boot tokens are then aggregated and refined by an \textbf{Aggregator} $f_{\text{merge}}$, leveraging the diversity and complementary strengths of different models while reducing error variance and mitigating early-stage hallucinations. This process repeats across layers, with only the top $k$ models generating further tokens conditioned on the intermediate answers. At each stage, the Selector re-evaluates outputs to minimize cumulative error $\epsilon_t$ and maintain semantic consistency.

This iterative, layer-wise routing and aggregation continues until the target output length is reached. Finally, the last-layer outputs are merged into a single response $Y^{\text{final}}$, balancing overall semantic consistency $S(Y^{\text{final}}, Y^{\text{true}})$ and total inference cost $T_\text{total}$.

\section{Experiments and Analysis}
\label{exp}
\subsection{Experimental Setup}

\paragraph{Datasets and Evaluation Metrics.}
We evaluate LightRouter on a range of widely-used benchmarks collectively referred to as \textbf{Knowledge and Reasoning Benchmarks}, covering diverse reasoning and knowledge-intensive tasks. Specifically, we include MMLU (5-shot)(~\cite{hendrycks2020measuring}) for general academic knowledge and reasoning (57 subjects), GSM8K (0-shot)(~\cite{cobbe2021training}) for grade-school math word problems, MATH (0-shot)(~\cite{hendrycks2021measuring}) for advanced mathematics, HumanEval(~\cite{chen2021evaluating}) for Python code generation, and GPQA-Diamond(~\cite{rein2024gpqa}) for graduate-level scientific reasoning in biology, physics, and chemistry. All experiments are conducted following official OpenCompass evaluation protocols. For multiple-choice tasks (MMLU, GPQA-Diamond), we report accuracy; for open-ended math (GSM8K, MATH), we use exact match (EM) accuracy; and for code generation (HumanEval), we report Pass@1, measuring the fraction of problems solved correctly on the first attempt.

Additionally, we report results on \textbf{MT-Bench} (~\cite{zheng2023judging}), where model responses are evaluated by GPT-4o in a pairwise preference scoring protocol.

\paragraph{Baselines}
To comprehensively validate the effectiveness of our framework, we consider the following baselines for comparison:
\begin{itemize}
    \item \textbf{Candidate Models:} Each individual base model incorporated in our framework is evaluated as a single-model baseline.
    \item \textbf{Comparative Models:} We include strong proprietary models such as \textit{deepseek-R1} and \textit{OpenAI o3-mini}, which demonstrate SOTA performance but incur higher inference costs.
    \item \textbf{Comparative Methods:} We compare with two representative multi-model aggregation strategies: \textit{MoA}(~\cite{wang2024mixture}), which uses a hierarchical composition of specialized models(with layer=1), and \textit{LLM-Blender}(~\cite{jiang2023llm}), which selects the best response using the PAIRRANKER reward model (with $K=1$). 
\end{itemize}

\paragraph{Implementation Details}
Our unified framework is implemented entirely with open-source models to ensure transparency and cost-effectiveness. The candidate model pool includes LLaMA-3.1-70B-Instruct (~\cite{touvron2023llama}), Mixtral-8x22B-v0.1 (~\cite{jiang2024mixtral}) from TogetherAI, Deepseek-v3 (~\cite{liu2024deepseek}), Qwen2.5-max, and Qwen2.5-Math-72B-Instruct (~\cite{qwen25}) from Alibaba's BAILIAN platform applicationcenter. Within LightRouter, \textbf{Deepseek-v3} serves as both the Selector and Aggregator module.

We evaluate two variants of LightRouter, with the primary setting employing a top-2 selection mechanism and a two-layer hierarchical routing strategy to boost output quality. All models are used strictly in accordance with their open-source licenses.

Following prior works (~\cite{yao2023tree,bian2023chatgpt,du2023improving}), we conduct experiments on subsets of the MATH and MMLU benchmarks. For fair comparison, all baseline and comparative methods are configured with the same set of candidate models as LightRouter. In particular, for the MoA baseline, Deepseek-v3 is used as the final proposer model and aggregator model to ensure consistency.

\subsection{Experiment Results}
\paragraph{Knowledge and Reasoning Benchmarks Performance}
\begin{table*}[htbp]
\centering
\footnotesize
\setlength{\tabcolsep}{4pt} 
\renewcommand{\arraystretch}{1.2}
\caption{\textbf{Knowledge and Reasoning Benchmarks results.} The best individual model is highlighted in red, and the best comparative method is highlighted in green. Our method is listed at the bottom and is presented in \textbf{bold} for emphasis.}
\resizebox{\textwidth}{!}{%
\begin{tabular}{lccccc}
\toprule
\textbf{Models} & \textbf{GSM8K/0-shot} & \textbf{MATH/0-shot} & \textbf{GPQA Diamond} & \textbf{HumanEval/pass@1} & \textbf{MMLU} \\
\midrule
\multicolumn{6}{c}{\textit{Candidate Models}} \\
Llama-3.1-70B-Instruct      & 92.68 & 67.80 & 48.01 & 80.50 & 86.00 \\
qwen2.5-max    & \cellcolor{pink!30}94.50 & 68.50 & 60.10 & 73.20 & 87.90 \\
qwen-2.5-Math-72B     & 90.80 & 88.10 & 39.39 & 70.73 & 81.13 \\
Mixtral-8x22B Instruct v0.1    & 83.70 & 41.80 & 34.30 & 46.30 & 77.75 \\
deepseek-v3      & 89.30 & 90.03 & 68.40 & 87.81 & 81.20 \\
\multicolumn{6}{c}{\textit{Comparative Models}} \\
deepseek-R1           & 93.40 & 97.30 & 71.50 & 95.73 & \cellcolor{pink!30}90.80 \\
OpenAI o3-mini           & 93.17 & \cellcolor{pink!30}97.90 & \cellcolor{pink!30}76.70 & \cellcolor{pink!30}97.60 & 86.90 \\
\midrule
\multicolumn{6}{c}{\textit{Comparative methods}} \\
LLM-BLENDER($K=1$)     & 94.30 & 83.00 & \cellcolor{green!15}67.60 & 74.39 & 87.52 \\
MoA           & \cellcolor{green!15}96.19 & \cellcolor{green!15}94.00 & 58.59 & \cellcolor{green!15}77.74 & \cellcolor{green!15}90.73 \\
\midrule
\multicolumn{6}{c}{\textit{Ours}} \\
\textbf{LightRouter}          & \textbf{97.196} & \textbf{94.30} & \textbf{73.34} & \textbf{94.43} & \textbf{91.56} \\
\bottomrule
\end{tabular}%
}
\label{tab:main_results}
\end{table*}

As summarized in Table~\ref{tab:main_results}, \textbf{LightRouter} consistently surpasses both strong individual models and advanced composite baselines across all benchmarks, achieving SOTA results: 97.20 on GSM8K, 94.30 on MATH, 73.34 on GPQA-Diamond, 94.43 on HumanEval, and 91.56 on MMLU. These results demonstrate that effective model selection and aggregation can substantially enhance overall performance. Compared to strong individual LLMs such as OpenAI o3-mini and Deepseek-R1, LightRouter consistently delivers superior results. Moreover, it outperforms advanced composite methods including \textbf{LLM-Blender} and \textbf{MoA}, achieving significant improvements across most benchmarks. These findings underscore the robustness, adaptability, and strong generalization capability of LightRouter.

\paragraph{Performance on MT-Bench}

\begin{figure}[htbp]
    \centering
    \includegraphics[width=1\linewidth]{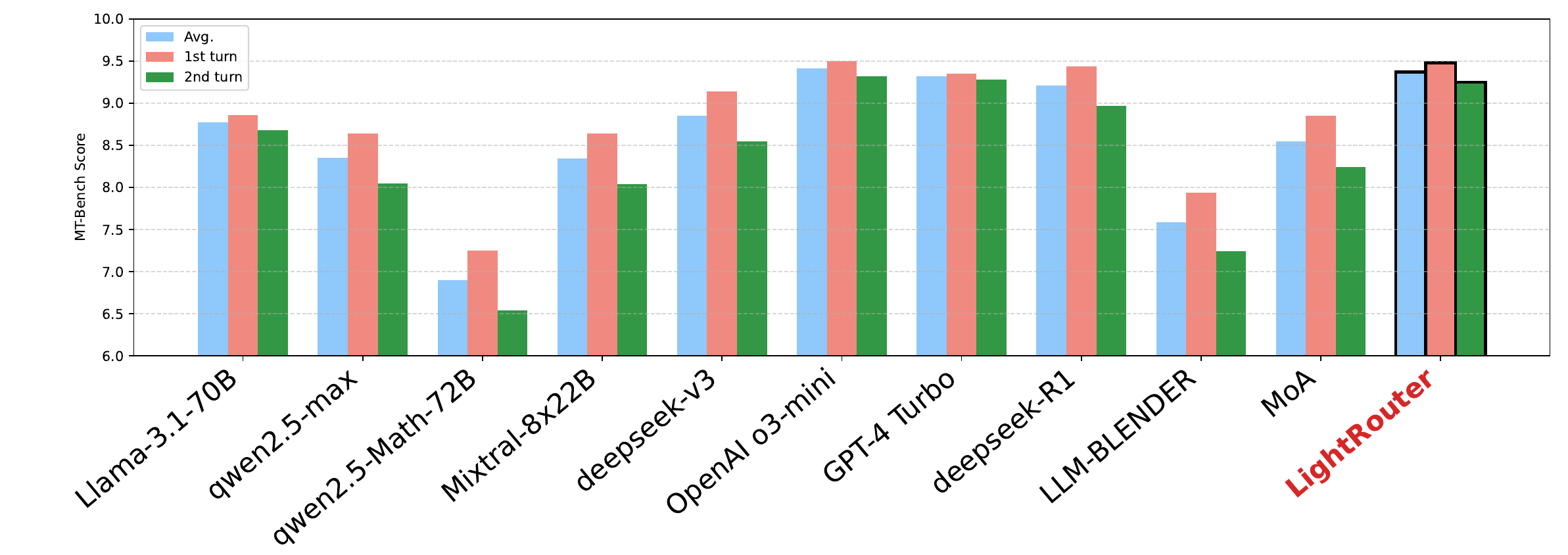}
    \caption{Grouped bar chart of MT-Bench scores for all evaluated models and methods. Each group represents a model, with bars indicating the average, 1st turn, and 2nd turn scores. LightRouter is highlighted in red and with a bold outline for emphasis.}
    \label{fig:mtbench_results}
\end{figure}

Figure~\ref{fig:mtbench_results} presents the MT-Bench evaluation results for all compared models and methods. \textbf{LightRouter} achieves an average score of 9.37, which is comparable to leading proprietary models such as OpenAI o3-mini (9.41), GPT-4 Turbo (9.32), and deepseek-R1 (9.21). Notably, LightRouter outperforms all candidate models and advanced composite methods, including LLM-BLENDER ($K=1$, 7.59) and MoA (8.55), across all metrics. In addition, LightRouter achieves the highest scores among all evaluated methods on both the first turn (9.48) and the second turn (9.25), indicating strong coherence and informativeness throughout multi-turn interactions. Although the absolute gains are modest due to the saturated performance of top models on MT-Bench, LightRouter consistently ranks among the best, highlighting the effectiveness and robustness of our routing framework in producing high-quality, human-aligned responses.

\section{Discussion}
\subsection{A Detailed Analysis on \textsc{LightRouter}}

To better understand the source of \textsc{LightRouter}'s effectiveness, we conduct a thorough analysis based on empirical evidence. Several key insights emerge from our results, which highlight the core design principles contributing to its superior performance.
\paragraph{Performance Across Benchmarks.}
First, \textsc{LightRouter} consistently outperforms all individual foundation models across a wide range of benchmarks, with particularly strong improvements on complex reasoning tasks such as \textbf{MATH} and \textbf{GSM8K}. This demonstrates the benefit of selective collaboration among large language models (LLMs), where queries are routed to the most appropriate model for each instance, as opposed to relying on the fixed capabilities of a single model.

\paragraph{Generalization Across Task Formats}
Furthermore, \textsc{LightRouter} generalizes well across diverse task formats, including multiple-choice QA, open-ended generation, and code synthesis, without requiring any task-specific tuning. This indicates the versatility of our framework in handling heterogeneous data modalities and evaluation protocols.

\paragraph{Feasibility with Open-Source Models}
Finally, it is worth emphasizing that all experimental results are obtained using only open-source LLMs. Despite this, \textsc{LightRouter} achieves performance that rivals or even surpasses proprietary APIs such as GPT-4. This affirms the feasibility of building competitive, cost-effective, and accessible multi-agent LLM systems entirely with open models.



\subsection{Why \textsc{LightRouter} Needs to Filter?}


\begin{figure}[htbp]
    \centering
    \includegraphics[width=1\linewidth]{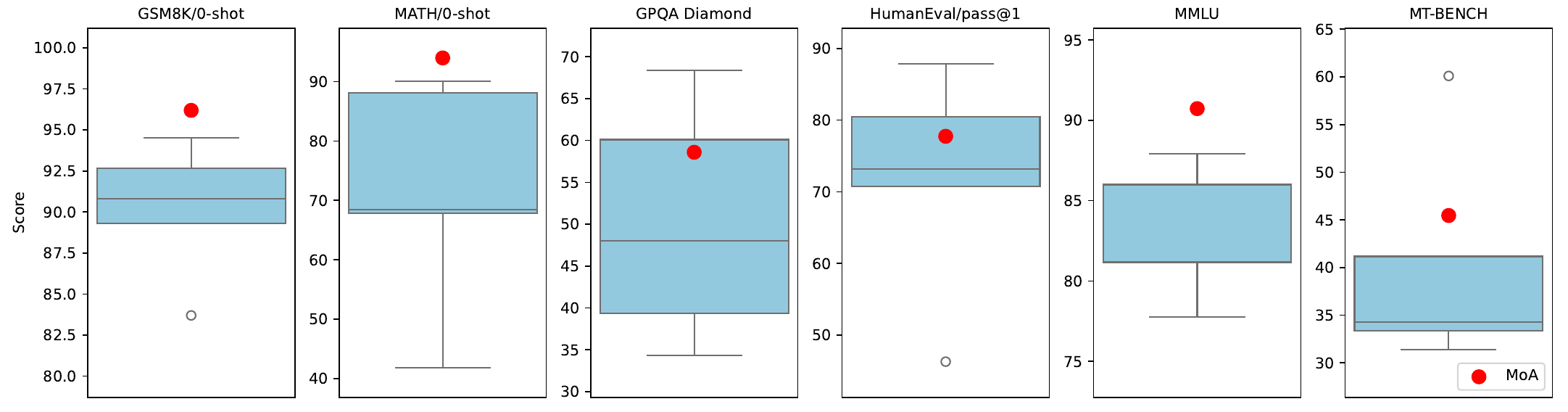}
    \caption{Boxplot comparison of individual candidate scores and aggregated MoA score across six benchmarks. Outliers are shown as small circles. The red dot denotes the MoA score for each task.}
    \label{fig:moa_box}
\end{figure}
Figure~\ref{fig:moa_box} illustrates the performance of the MoA method when the candidate pool includes models of varying quality. Our results challenge the common assumption that low-quality models will always undermine aggregation: the actual impact depends heavily on the distribution of model performances. In some cases, introducing a few poor-performing models can severely degrade the aggregated results, while in others, the effect is more moderate. This variability highlights that MoA’s effectiveness is not guaranteed and is highly sensitive to the composition of the candidate pool. Retaining low-quality responses increases the risk of substantial performance drops, indicating that effective filtering is essential. Removing poorly performing models or outputs helps ensure that aggregation remains robust, reducing the likelihood of outcome deterioration due to outliers.

\subsection{Why \textsc{LightRouter} Needs to Aggregate?}

Previous methods often focus on selecting the "best" single model for each query or routing tasks to different experts based on static preferences. However, such approaches overlook a critical challenge: even SOTA LLMs exhibit significant output variability when presented with the same prompt multiple times.

\begin{figure}
    \centering
    \includegraphics[width=1\linewidth]{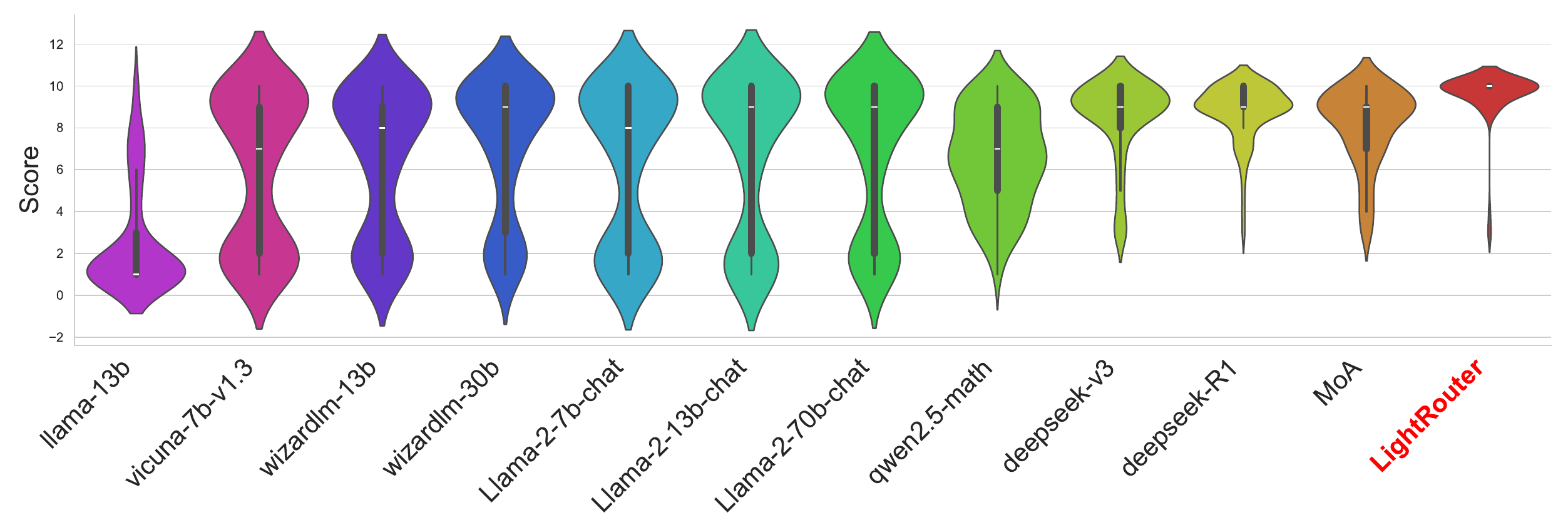}
    \caption{Performance distribution of different LLMs across repeated runs on MT-Bench. Our method (\textsc{LightRouter}) shows both high average performance and low variance.}
    \label{fig:violin}
\end{figure}

To empirically illustrate this phenomenon, we conduct an experiment measuring the performance distribution of several LLMs on repeated runs of the same MT-Bench queries. As shown in Figure~\ref{fig:violin}, the performance of individual models can vary widely across different samples, with substantial overlap in their score distributions. This variability suggests that relying on a single model or a single response is insufficient for robust and reliable question answering. To further investigate, we also include ensemble-based approaches such as MoA and our proposed LightRouter in the analysis. The results show that both MoA and LightRouter exhibit significantly reduced variance in their score distributions compared to individual candidate models.


\begin{wrapfigure}{r}{0.40\textwidth}
  \vspace{-18pt} 
  \centering
  \includegraphics[width=0.36\textwidth]{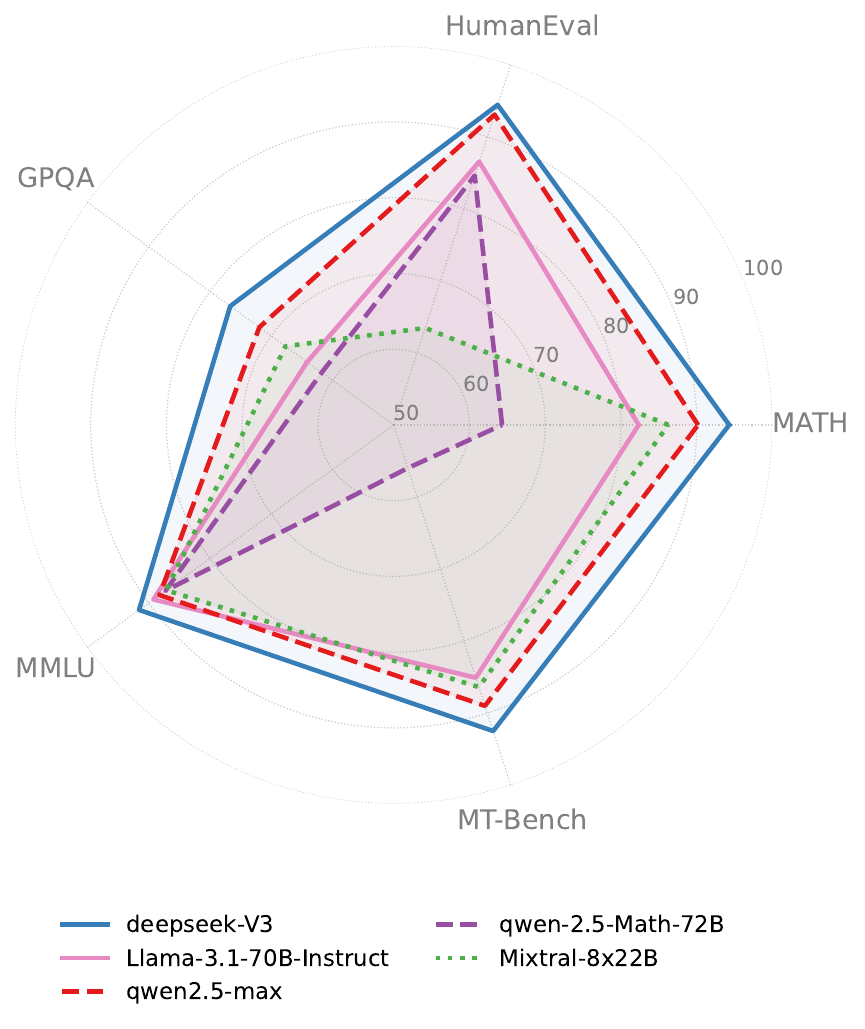}
  \caption{Performance of different LLMs when used as aggregators.}
  \label{fig:radar_chart}
  \vspace{-35pt} 
\end{wrapfigure}
Moreover, to systematically evaluate the impact of the aggregator, we select a diverse set of open-source LLMs with varying capacities and instruction-following abilities to serve as aggregators. Specifically, we include both lightweight models (e.g., Llama-3.1-8B-Instruct) and larger, more advanced options (e.g., Deepseek-v3, qwen2.5-max, Mixtral-8x22B). All experiments are conducted under the same candidate input settings to ensure fairness.

As shown in Figure ~\ref{fig:radar_chart}, the choice of aggregator has a significant effect on final performance. Overall, these results demonstrate that the aggregator's capacity and training objectives greatly influence the quality of the merged outputs. Stronger aggregators are better at synthesizing and selecting among candidate responses, leading to more reliable final answers in the multi-model framework.

\begin{figure}[htbp]
    \centering
    \begin{minipage}[t]{0.49\textwidth}
        \centering
        \includegraphics[height=3.8cm]{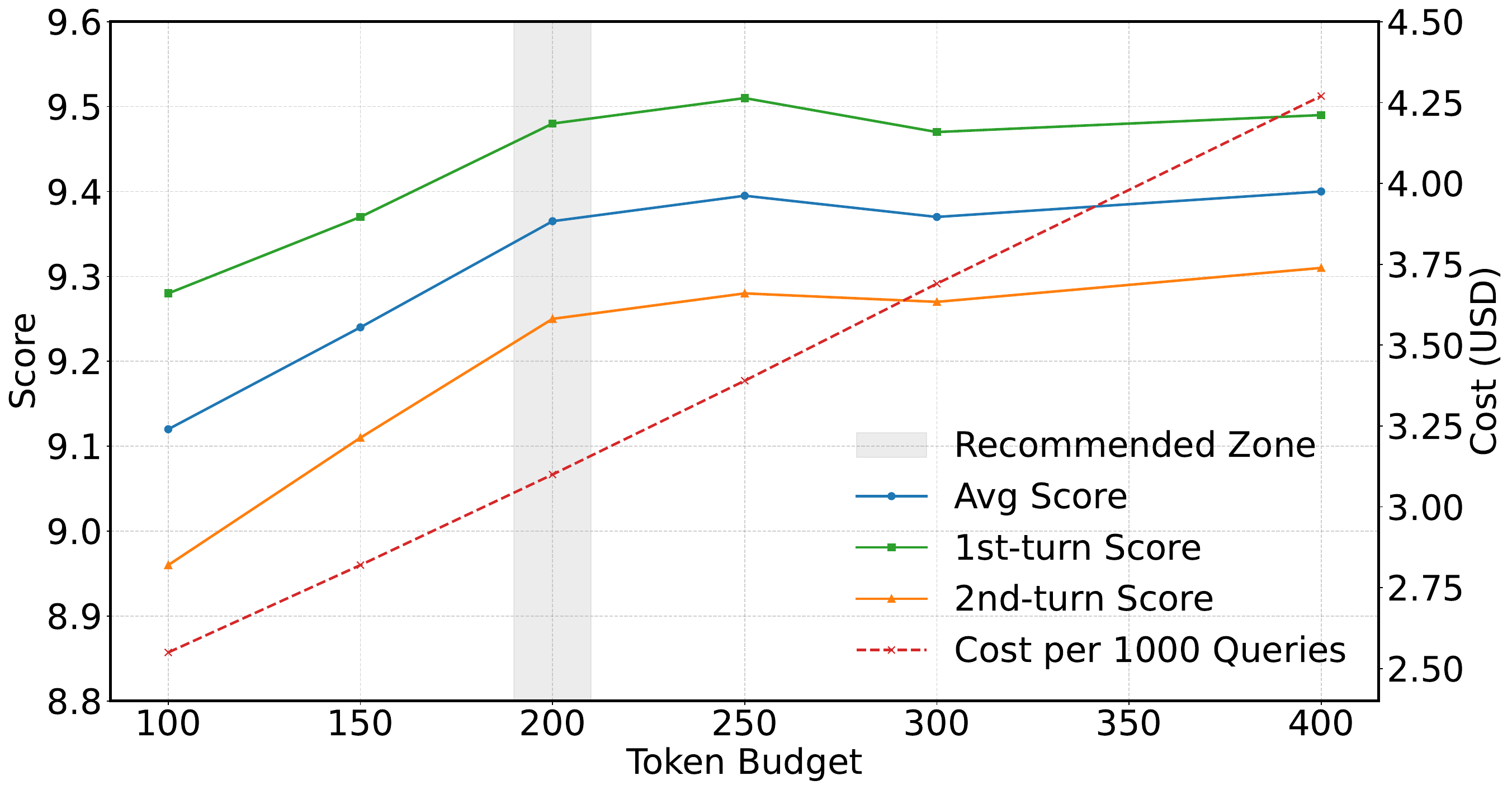}
        \caption{\textbf{Trade-off analysis between token budget, accuracy, and cost.}
        Accuracy increases marginally with more tokens, while cost rises steadily.
        The 200-token point provides a practical balance.}        
        \label{fig:token_cost}
    \end{minipage}%
    \hfill
    \begin{minipage}[t]{0.47\textwidth}
        \centering
        \includegraphics[height=3.8cm]{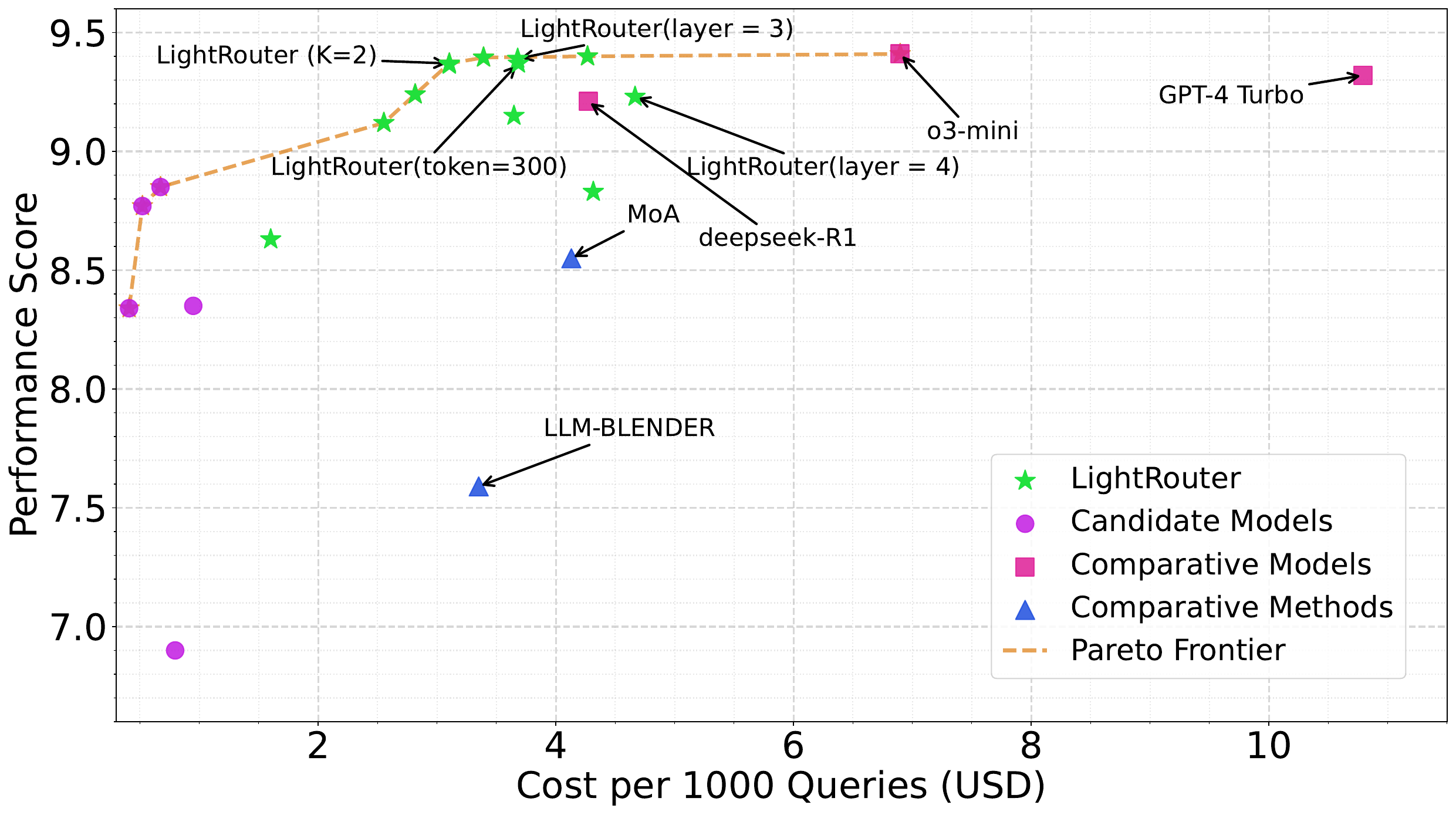}
        \caption{\textbf{Performance-cost tradeoff of LightRouter versus baselines.} Across different configurations, LightRouter consistently demonstrates superior performance-cost efficiency, as evidenced by its placement on the Pareto frontier.}
        \label{fig:pareto_front}
    \end{minipage}
\end{figure}
\subsection{How Will Boot Token Numbers Influence?}
This experiment systematically investigates the trade-off between token budget, accuracy, and inference cost. As shown in Figure~\ref{fig:token_cost}, increasing the token budget leads to a consistent improvement in both the average score and the scores for the first and second turns. However, the rate of improvement diminishes as the token count increases: the gain in accuracy is substantial when raising the budget from 100 to 200 tokens, but further increases yield only marginal benefits, with some scores even exhibiting slight fluctuations. Taken together, these results indicate that the 200-token budget represents an effective balance point: the model achieves near-optimal accuracy, while the associated cost remains moderate. Thus, selecting a 200-token budget for deployment offers a practical compromise between performance and cost.


With expanded experimental settings, we systematically evaluate \textsc{LightRouter} across three key dimensions: (1) varying the number of selected models ($k=1$ to $4$) with a fixed token budget of 200 and two layers; (2) varying the token budget from 100 to 400 with $k=2$ and two layers; and (3) varying the number of layers ($l=2$ to $4$) with $k=2$ and a fixed token budget of 200. This comprehensive analysis enables a thorough examination of LightRouter’s cost and performance trade-offs.

Table~\ref{tab:llm_statistics} presents the unit API costs of all models, and Figure~\ref{fig:pareto_front} visualizes the performance-cost landscape for all configurations.

Compared to ensemble-based baselines such as MoA and LLM-Blender, LightRouter achieves superior accuracy while significantly reducing computational cost. This is enabled by our \textbf{token-efficient routing mechanism}, which makes early routing decisions based on partial outputs (e.g., the first 200 tokens), avoiding the high latency and cost of generating full responses from all models. Quantitatively, LightRouter reduces API costs per query by over 7.46\% compared to LLM-Blender and by 24.85\% compared to MoA, with even greater cost advantages as the number of candidate LLMs increases.

We further examine the impact of multi-layer aggregation by varying the number of layers from 2 to 4 (with $k=2$). As the number of layers increases, we observe that the average score improves slightly from 9.37 (cost: 0.0031) at $l=2$ to 9.39 (cost: 0.0037) at $l=3$, but then declines to 9.23 (cost: 0.0047) at $l=4$. This indicates that while deeper aggregation can offer marginal gains, increasing the number of layers beyond three results in diminishing returns and higher inference cost.

LightRouter also outperforms SOTA single models with strong reasoning abilities, such as DeepSeek-R1, GPT-4o, and GPT-o3 mini, reducing per-query costs by 27.34\%, 71.24\%, and 54.99\% respectively, while maintaining comparable accuracy.

Finally, using \textbf{Pareto Optimality} to analyze the performance-cost trade-off, Figure~\ref{fig:pareto_front} shows that LightRouter consistently lies on the Pareto frontier, demonstrating its effectiveness as a cost-efficient solution for practical deployment.

\section{Related Work}
\subsection{Model Selection}
Online model selection aims to identify the best candidate model efficiently(~\cite{foster2017parameter, foster2019model, ravaut2022summareranker, liu2021simcls, owodunni2023koya, ouyang2022training}). While traditional methods and reranking (e.g., LLM-Blender~\cite{jiang2023llm}) can be effective, they require all models to produce full outputs, incurring high computational cost. To address scalability, router-based approaches (~\cite{wang2023fusing, shnitzer2023large, lu2023routing}) predict the most promising LLM for each input, but their performance depends on router accuracy and adaptability. Recent advances such as GraphRouter(~\cite{feng2024graphrouter}) and TI-UCB(~\cite{xia2024llm}) further improve efficiency and adaptability in model selection, while RouterEval(~\cite{huang2025routereval}) provides comprehensive benchmarks for evaluating router-based methods.

\subsection{Model Ensemble}
Several works seek to harness the synergy of multiple LLMs by aggregating their outputs. GENFUSER(~\cite{jiang2023llm}) trains a fusion model to integrate candidate answers, but such training-dependent methods often struggle to generalize across diverse tasks or domains. In contrast, Huang et al.(~\cite{huang2024ensemble}) propose a theoretical framework for ensembling heterogeneous LLMs by aligning outputs in a shared embedding space. To address vocabulary mismatches, MINED(~\cite{wan2024knowledge,fu2023specializing}) aligns tokens based on edit distance, enabling joint reasoning. ReConcile(~\cite{chen2023reconcile}) models multi-agent communication via round-table discussion and consensus voting. Another line of work draws inspiration from Mixture-of-Experts (MoE) architectures(~\cite{jacobs1991adaptive,shazeer2017outrageously,du2022glam}), such as MoA(~\cite{wang2024mixture}), which assigns specialized roles (e.g., proposers, aggregators) to different models for hierarchical collaboration. While these approaches improve performance through collaboration, they typically require querying all models, leading to substantial token consumption and computational overhead, thus limiting practical scalability.


\section{Conclusion}

We present \textsc{LightRouter}, a token-efficient routing framework that addresses the major challenges of integrating multiple LLMs—namely, high inference cost and inefficient model coordination. By making early routing decisions based on partial outputs and selectively aggregating the most promising responses, \textsc{LightRouter} achieves strong performance across a variety of knowledge and reasoning benchmarks, while substantially reducing computational overhead.

Our experiments show that \textsc{LightRouter} not only outperforms strong individual LLMs but also surpasses advanced ensemble methods such as LLM-Blender and MoA in both accuracy and cost-effectiveness. Its scalable design and ability to filter out noisy completions make it robust to the variability of modern LLM outputs. Importantly, \textsc{LightRouter} achieves these gains using only open-source models, demonstrating the feasibility of building powerful and accessible LLM ensembles without reliance on proprietary APIs.

Looking ahead, further research on adaptive scoring and dynamic specialization could enhance coordination and efficiency, while extending \textsc{LightRouter} to multi-modal or continual learning contexts remains a promising direction. Overall, \textsc{LightRouter} offers an efficient, reliable, and scalable solution for LLM orchestration, paving the way for broader adoption of collaborative language model systems.

\section{Limitation}
\label{limitation}
LightRouter, while effective on standard benchmarks, has several limitations. First, its adaptability to open-domain and novel tasks remains uncertain, as its performance on out-of-distribution or few-shot problems has not been fully explored. Second, the current framework only supports selection among non-reasoning models. For reasoning-intensive tasks where most inference cost comes from CoT steps, LightRouter cannot limit CoT length due to lack of API support for truncated CoT outputs. Third, the method uses static, uniform token budgets and selection stages for all queries, lacking dynamic adjustment based on input difficulty or type, which may reduce efficiency or accuracy in real-world applications.




\newpage
\bibliographystyle{plainnat} 

\bibliography{ref}   
\newpage
\appendix
\section{Model API Prices}

\begin{table}[htbp]
\centering
\footnotesize
\setlength{\tabcolsep}{12pt} 
\renewcommand{\arraystretch}{1.2} 
\caption{Statistics of Different LLMs and Their API Costs.}
\label{tab:llm_statistics}
\begin{tabular}{l c}
\toprule
\textbf{LLM} & \textbf{Cost per 1M tokens} \\
\midrule
Llama-3.1-70B-Instruct & 0.88 \\
qwen2.5-max & 1.32 \\
qwen-2.5-Math-72B & 1.65 \\
Mixtral-8x22B Instruct v0.1 & 0.80 \\
deepseek-v3 & 1.10 \\
deepseek-R1 & 2.20 \\
OpenAI o3-mini & 4.40 \\
GPT-4 Turbo (04/09) & 30\\
\bottomrule
\end{tabular}
\end{table}

\section{Prompts for LightRouter}
\begin{table}[htbp]
\centering
\begin{minipage}{0.95\textwidth}
\renewcommand{\arraystretch}{1.2}
\caption{Aggregate-and-Synthesize Prompt to integrate responses from other models.}
\label{tab:agg-synth-prompt}
\begin{tabular}{|p{0.96\textwidth}|}
\hline
You are given a set of responses from various open-source models to a user query. Your task is to synthesize these into a single, high-quality response.

Critically evaluate the input responses — some may contain biases, errors, or irrelevant content. Do not simply merge or repeat them. Instead, produce a refined, accurate, and comprehensive answer that reflects the best possible understanding of the query.

Your final response must be well-structured, coherent, and meet the highest standards of accuracy and clarity.

Input model responses:

LLM1's answer: \{answer1\}

LLM2's answer: \{answer2\}

query: \{state["query"]\}
\\
\hline
\end{tabular}
\end{minipage}
\end{table}

\begin{table}[htbp]
\centering
\begin{minipage}{0.95\textwidth}
\renewcommand{\arraystretch}{1.2}
\caption{Aggregate-and-Synthesize Prompt to integrate responses from other models.}
\label{tab:agg-synth-prompt}
\begin{tabular}{|p{0.96\textwidth}|}
\hline
You are an intelligent agent skilled at predicting the overall quality of a complete LLM response based on partial outputs.

You will be given a query and partial responses from five different LLMs. Your task is to rank the LLMs based on the predicted quality of their full responses.

Instructions:

1.Generate your own brief answer to the query to help assess logic and accuracy (do not assume it is always correct).

2.Analyze each partial response and evaluate the logical correctness and clarity of reasoning.

3.Rank the LLMs by how likely their complete answers would be high-quality and accurate.

Output Format (Required):
Only output a list in this format:
[LLM1, LLM2, LLM3, LLM4, LLM5]
Order from highest to lowest predicted response quality.

Important:
DO NOT include any explanations, comments, or additional text — only return the ranking list.

query: {state["query"]}

LLM1: {answer1}

LLM2: {answer2}

LLM3: {answer3}

LLM4: {answer4}

LLM5: {answer5}
\\
\hline
\end{tabular}
\end{minipage}
\end{table}

\newpage
\section*{NeurIPS Paper Checklist}

\begin{enumerate}

\item {\bf Claims}
    \item[] Question: Do the main claims made in the abstract and introduction accurately reflect the paper's contributions and scope?
    \item[] Answer: \answerYes{}
    \item[] Justification: The main claims are clearly stated in the abstract and introduction and are consistent with the theoretical and experimental results presented in the paper.
    
\item {\bf Limitations}
    \item[] Question: Does the paper discuss the limitations of the work performed by the authors?
    \item[] Answer: \answerYes{}
    \item[] Justification: The paper includes a dedicated "Limitations" section~\ref{limitation} discussing the main assumptions, robustness to violations, and the scope of applicability of our method.

\item {\bf Theory assumptions and proofs}
    \item[] Question: For each theoretical result, does the paper provide the full set of assumptions and a complete (and correct) proof?
    \item[] Answer: \answerYes{}
    \item[] Justification: All theoretical results are accompanied by clearly stated assumptions and complete proofs.

\item {\bf Experimental result reproducibility}
    \item[] Question: Does the paper fully disclose all the information needed to reproduce the main experimental results of the paper to the extent that it affects the main claims and/or conclusions of the paper (regardless of whether the code and data are provided or not)?
    \item[] Answer: \answerYes{}
    \item[] Justification: All experimental settings, hyperparameters, and procedures are described in Section \ref{exp}.

\item {\bf Open access to data and code}
    \item[] Question: Does the paper provide open access to the data and code, with sufficient instructions to faithfully reproduce the main experimental results, as described in supplemental material?
    \item[] Answer: \answerYes{}
    \item[] Justification: An anonymized GitHub repository containing code, data, and a detailed README with instructions to reproduce all main results is provided in the supplemental material.

\item {\bf Experimental setting/details}
    \item[] Question: Does the paper specify all the training and test details (e.g., data splits, hyperparameters, how they were chosen, type of optimizer, etc.) necessary to understand the results?
    \item[] Answer: \answerYes{}
    \item[] Justification: Section \ref{exp} provide all relevant experimental details, including dataset splits, hyperparameters, selection methods, and computational environment.

\item {\bf Experiment statistical significance}
    \item[] Question: Does the paper report error bars suitably and correctly defined or other appropriate information about the statistical significance of the experiments?
    \item[] Answer: \answerYes{}
    \item[] Justification: Justification: The main results are reported as averages in Table \ref{tab:main_results}. To further illustrate the stability and statistical significance, we present a violin plot (Figure \ref{fig:violin}), which depicts the distribution, mean, and variance of scores for each model. The violin plot provides a comprehensive view of the variability and robustness of our results..

\item {\bf Experiments compute resources}
    \item[] Question: For each experiment, does the paper provide sufficient information on the computer resources (type of compute workers, memory, time of execution) needed to reproduce the experiments?
    \item[] Answer: \answerYes{}
    \item[] Justification: All calculation is done with api call. Cloud provider is illustrated in experiment settings.

\item {\bf Code of ethics}
    \item[] Question: Does the research conducted in the paper conform, in every respect, with the NeurIPS Code of Ethics \url{https://neurips.cc/public/EthicsGuidelines}?
    \item[] Answer: \answerYes{}
    \item[] Justification: All research procedures comply with the NeurIPS Code of Ethics, including fair data usage, privacy, and societal impact considerations .

\item {\bf Broader impacts}
    \item[] Question: Does the paper discuss both potential positive societal impacts and negative societal impacts of the work performed?
    \item[] Answer: \answerYes{}
    \item[] Justification: While our work is foundational, we note in the Introduction and Conclusion that more efficient LLM integration could lower barriers for misuse, such as generating disinformation or amplifying biases. We recommend responsible deployment and continuous monitoring to mitigate these risks (see Introduction and Conclusion for further discussion).

\item {\bf Safeguards}
    \item[] Question: Does the paper describe safeguards that have been put in place for responsible release of data or models that have a high risk for misuse (e.g., pretrained language models, image generators, or scraped datasets)?
    \item[] Answer: \answerNA{}
    \item[] Justification: The paper does not release any model or data with high risk for misuse.

\item {\bf Licenses for existing assets}
    \item[] Question: Are the creators or original owners of assets (e.g., code, data, models), used in the paper, properly credited and are the license and terms of use explicitly mentioned and properly respected?
    \item[] Answer: \answerYes{}
    \item[] Justification: All third-party assets are properly cited.

\item {\bf New assets}
    \item[] Question: Are new assets introduced in the paper well documented and is the documentation provided alongside the assets?
    \item[] Answer: \answerYes{}
    \item[] Justification: The released dataset and code are  included in the supplementary material.

\item {\bf Crowdsourcing and research with human subjects}
    \item[] Question: For crowdsourcing experiments and research with human subjects, does the paper include the full text of instructions given to participants and screenshots, if applicable, as well as details about compensation (if any)?
    \item[] Answer: \answerNA{}
    \item[] Justification: This paper does not involve crowdsourcing or research with human subjects.

\item {\bf Institutional review board (IRB) approvals or equivalent for research with human subjects}
    \item[] Question: Does the paper describe potential risks incurred by study participants, whether such risks were disclosed to the subjects, and whether Institutional Review Board (IRB) approvals (or an equivalent approval/review based on the requirements of your country or institution) were obtained?
    \item[] Answer: \answerNA{}
    \item[] Justification: No human subjects were involved in this research.

\item {\bf Declaration of LLM usage}
    \item[] Question: Does the paper describe the usage of LLMs if it is an important, original, or non-standard component of the core methods in this research? Note that if the LLM is used only for writing, editing, or formatting purposes and does not impact the core methodology, scientific rigorousness, or originality of the research, declaration is not required.
    \item[] Answer: \answerNA{}
    \item[] Justification: LLMs were used only for language editing and writing assistance, not as an important, original, or non-standard component of the core methods. According to the NeurIPS LLM policy, such usage does not require declaration.

\end{enumerate}

\end{document}